\documentclass[10pt,letterpaper]{article}
\usepackage[top=0.85in,left=2.75in,footskip=0.75in]{geometry}

\usepackage{siunitx}
\usepackage{pdfpages}

\usepackage{amsmath,amssymb}

\usepackage{changepage}

\usepackage[utf8x]{inputenc}

\usepackage{textcomp,marvosym}

\usepackage{cite}

\usepackage{nameref,hyperref}

\usepackage[right]{lineno}

\usepackage{microtype}
\DisableLigatures[f]{encoding = *, family = * }

\usepackage{array}

\newcolumntype{+}{!{\vrule width 2pt}}

\newlength\savedwidth

\usepackage{setspace} 

\usepackage[aboveskip=3pt,labelfont=bf,labelsep=period,justification=raggedright,singlelinecheck=off]{caption}

\makeatletter
\renewcommand{\@biblabel}[1]{\quad#1.}
\makeatother

\hypersetup{
	linkcolor=blue!50!black!100,
	colorlinks=true,
	citecolor=blue!50!black!100,
	urlcolor=blue!50!black!100,
}

\usepackage{lastpage,fancyhdr,graphicx}
\usepackage{epstopdf}
\fancyheadoffset[L]{2.25in}
\fancyfootoffset[L]{2.25in}
\begin{document}
\vspace*{0.2in}

\begin{flushleft}
{\Large
\textbf\newline{Cell segmentation and tracking using CNN-based distance predictions and a graph-based matching strategy} 
}
\newline
\\
Tim Scherr\textsuperscript{1*},
Katharina Löffler\textsuperscript{1,2},
Moritz Böhland\textsuperscript{1},
Ralf Mikut\textsuperscript{1}
\\
\bigskip
\textbf{1} Institute for Automation and Applied Informatics, Karlsruhe Institute of Technology, Eggenstein-Leopoldshafen, Germany
\\
\textbf{2} Institute of Biological and Chemical Systems - Biological Information Processing, Karlsruhe Institute of Technology, Eggenstein-Leopoldshafen, Germany
\\
\bigskip

%
%





* tim.scherr@kit.edu

\end{flushleft}
\section*{Abstract}
The accurate segmentation and tracking of cells in microscopy image sequences is an important task in biomedical research, e.g., for studying the development of tissues, organs or entire organisms. However, the segmentation of touching cells in images with a low signal-to-noise-ratio is still a challenging problem. In this paper, we present a method for the segmentation of touching cells in microscopy images. By using a novel representation of cell borders, inspired by distance maps, our method is capable to utilize not only touching cells but also close cells in the training process. Furthermore, this representation is notably robust to annotation errors and shows promising results for the segmentation of microscopy images containing in the training data underrepresented or not included cell types. For the prediction of the proposed neighbor distances, an adapted U-Net convolutional neural network~(CNN) with two decoder paths is used. In addition, we adapt a graph-based cell tracking algorithm to evaluate our proposed method on the task of cell tracking. The adapted tracking algorithm includes a movement estimation in the cost function to re-link tracks with missing segmentation masks over a short sequence of frames. Our combined tracking by detection method has proven its potential in the IEEE ISBI 2020 Cell Tracking Challenge (\href{http://celltrackingchallenge.net/}{http://celltrackingchallenge.net/}) where we achieved as team KIT-Sch-GE multiple top three rankings including two top performances using a single segmentation model for the diverse data sets.

\section*{Introduction}
\label{sec:introduction}State-of-the-art microscopy imaging techniques such as light-sheet fluorescence microscopy imaging enable to investigate cell dynamics with single-cell resolution~\cite{Chhetri2015, Kobitski2015}. This allows to study cell migration and proliferation in tissue development and organ formation at early embryonic stages. Establishing the required complete lineage of each cell, however, requires a virtually error-free segmentation and tracking of individual cells over time~\cite{Kobitski2015,Khairy2011}. A manual data analysis is unfeasible, due to the large amount of data acquired with modern imaging techniques. In addition, low-resolution objects are very difficult to detect even for human experts. Deep learning-based cell segmentation methods have proven to outperform traditional methods even on very diverse 2D data sets~\cite{Caicedo2019}. However, state-of-the-art cell tracking methods often still need a time-consuming manual cell track curation, e.g., using EmbryoMiner~\cite{Schott2018} or the Massive Multi-view Tracker (MaMuT)~\cite{Wolff2018}. Especially for low signal-to-noise ratio and 3D data, further method development is required for both cell segmentation and cell tracking~\cite{Ulman2017}.

Traditional segmentation methods, such as TWANG for the segmentation of roundish objects~\cite{Stegmaier2014}, are often designed for a specific application. These methods commonly consist of sophisticated combinations of pre-processing filters, e.g., Gaussian or median filters, and segmentation operations, e.g., a region adaptive thresholding followed by a watershed transform~\cite{Maska2014}. To reach a reasonable segmentation quality, such traditional methods need to be carefully adapted to the cell type and imaging conditions. Therefore, expert knowledge is needed. In contrast, deep learning-based segmentation methods shift the expert knowledge needed to the model design and to the training process. Thus, less expert knowledge is needed for the application of a trained model and to fine-tune the post-processing which is often kept very simple. A review of cell segmentation methods is provided in \cite{Vicar2019}.

\begin{figure}
\begin{adjustwidth}{-2.25in}{0in}
\centering
\includegraphics[scale=1]{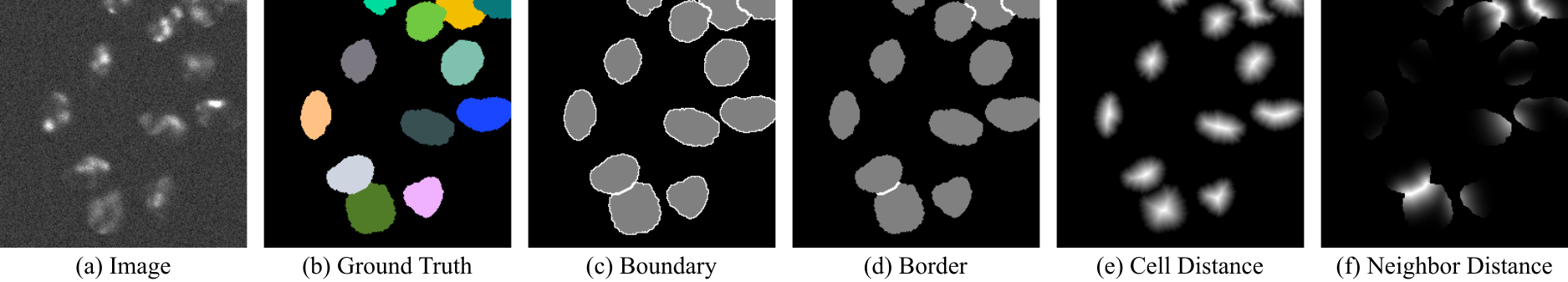}
\caption{\textbf{Training data representations for the training of deep learning models.} Image (a) and ground truth (b) show a crop of the simulated Cell Tracking Challenge data set Fluo-N2DH-SIM+~\cite{Ulman2017, Maska2014}. Generated boundaries (c) and borders (d) can be used to split touching cells. Many training data sets contain only few touching cells resulting in few training samples for borders and boundaries between cells. The combination of cell distances (e) with neighbor distances (f) is aimed to solve this problem since models can also learn from close cells.}
\label{fig:trainingDataRepresentations}
\end{adjustwidth}
\end{figure}

To improve the generalization ability of a trained deep learning model, a preferably diverse and large annotated data set is needed. This fact is especially problematic when dealing with touching cells since this case is usually underrepresented in training data sets. Therefore, models for cell boundary or border prediction (see Fig~\ref{fig:trainingDataRepresentations}) are often not able to handle touching cells well. The result are merged cells, due to gaps in predicted cell boundaries and borders between touching cells~\cite{Scherr2018, Pena2019}. To overcome this problem, several approaches have been proposed. In~\cite{Scherr2018}, models are trained to predict adapted thicker borders and smaller cells, which can decrease the amount of merged cells. \cite{Pena2019}~utilizes new gap and touching classes with $J$ regularization. \cite{Li2019}~combines distance transforms for single cell nuclei with discrete boundaries. A center vector encoding which is aimed to be more robust to label inconsistencies is proposed in~\cite{Li2019b}, whereas in~\cite{Graham2019}, horizontal and vertical gradient maps are used. To improve the generalization ability of a model for cell types with only few or no annotated images, a generative adversarial network-based image style transfer to generate augmented training samples of that cell types has been used in~\cite{Hollandi2019}. An advantage of border-based approaches is that a deep learning model is enforced to focus on touching cells that are underrepresented in the training data. However, border-based approaches still have the shortcoming that only touching cells can be used to train the border prediction.

Although deep learning methods have been successfully applied to multi object tracking on natural images~\cite{Ciaparrone2020, Yao2019}, there are only few deep learning approaches for cell tracking~\cite{Payer2019, He2017}. In~\cite{Payer2019}, cells are simultaneously segmented and tracked by combining a recurrent hourglass network with a pixel-wise metric embedding learning. \cite{He2017} proposes a particle-filter-based motion model in combination with a CNN-based observation model. However, cell tracking is still dominated by traditional tracking approaches~\cite{Ulman2017, Magnusson2015}. One reason is the lack of high quality annotations as provided in natural image tracking benchmarks~\cite{MOTS2019, DAVIS2017, YTVOS2018}. Thus, training data are often not available. Another aspect that complicates the task of cell tracking are cell death and division events, which do not occur in natural image tracking data. Therefore, traditional tracking algorithms with comparably few parameters and explicit modeling of cell division events still dominate cell tracking benchmarks~\cite{Ulman2017}. The comparison of cell tracking algorithms in~\cite{Ulman2017} shows that the majority of tracking approaches uses an adapted version of nearest neighbors, a graph-based linking or multi hypothesis tracking. In \cite{Magnusson2015}, the Viterbi algorithm is applied to track cells. A joint model for segmentation and tracking is proposed in \cite{Sixta2020} where model parameters are learned based on Bayes risk minimization.

In this paper, we propose a novel representation of cell borders, the neighbor distances, to solve the challenging problem of segmenting touching cells of various types in the absence of large training data sets. Thus far, problems of border prediction approaches are the sensitivity to annotation inconsistencies, and that only touching cells provide border information in the training. The neighbor distances are aimed to be less sensitive to annotation inconsistencies, and enable to learn also from close cells. This additional information in the training process results in a more robust border prediction. Similar to~\cite{Li2019}, we combine our border predictions with cell distances to further prevent the erroneous merging of close cells. However, in contrast to~\cite{Li2019}, we remove the bottlenecks of non-robust discrete boundaries and of the feature fusion layers. This results in a simplified architecture and training process and in less merged cells. For the cell tracking, we adapt a coupled minimum-cost flow algorithm to include an object movement estimation. In addition, our formulation is able to link fragmented tracks due to missing segmentation masks in a short sequence of frames. The remainder of this paper covers the methodology we use to detect and segment cells and the subsequent cell tracking. In the results section, we demonstrate the quality of our introduced method on data from the Cell Tracking Challenge \cite{Ulman2017, Maska2014}.

\section*{Materials and methods}
\subsection*{Cell segmentation using CNN-based distance predictions}
\begin{figure}
\begin{adjustwidth}{-2.25in}{0in}
\centering
\includegraphics[scale=1]{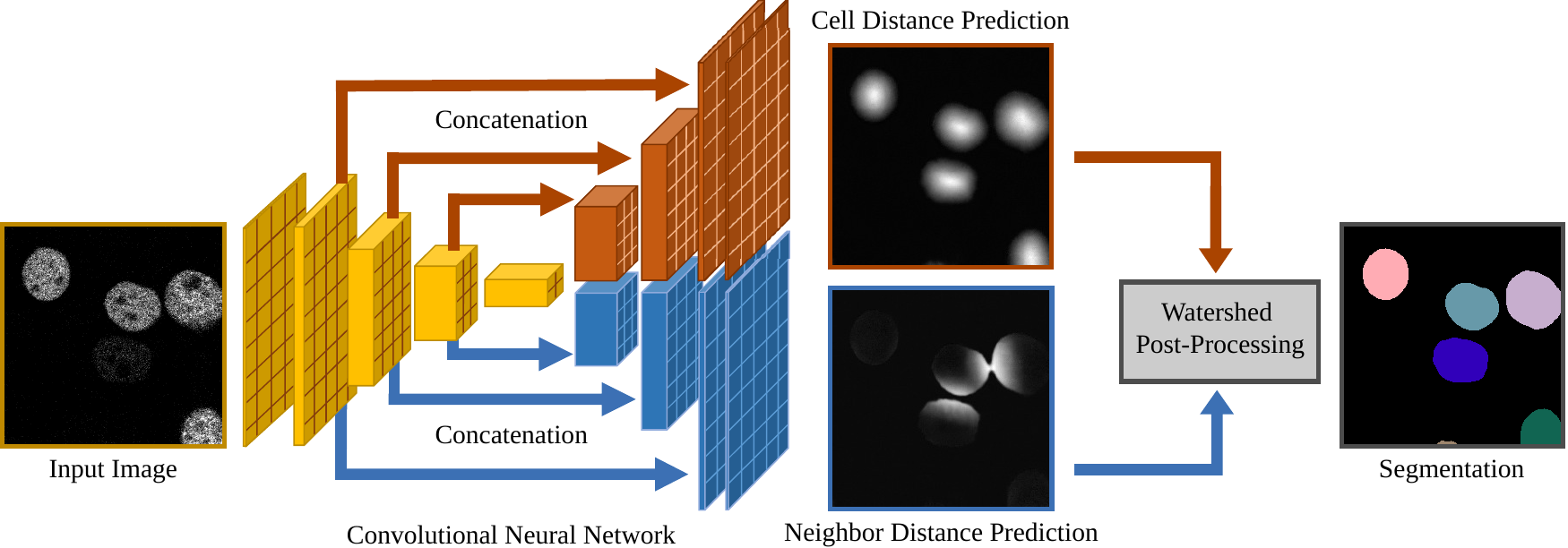}
\caption{\textbf{Overview of the proposed segmentation method using distance predictions} (adapted from~\cite{Li2019}). The CNN consists of a single encoder that is connected to both decoder paths. The network is trained to predict cell distances and neighbor distances that are used for the watershed-based post-processing. The input image shows a crop of the Cell Tracking Challenge data set Fluo-N2DH-GOWT1~\cite{Ulman2017, Maska2014}.}
\label{fig:architecture}
\end{adjustwidth}
\end{figure}

For cell segmentation, we train a deep learning model to predict cells and cell borders, followed by a post-processing with a seed extraction and a seed-based watershed segmentation. A key for the successful application of supervised deep learning methods in the absence of large training data sets is to introduce representations that allow to use as much information as possible. Thus, instead of discrete cell boundary (Fig~\ref{fig:trainingDataRepresentations}c) and cell border representations (Fig~\ref{fig:trainingDataRepresentations}d), we combine cell distances~(Fig~\ref{fig:trainingDataRepresentations}e, \cite{Li2019}) with novel neighbor distances (Fig~\ref{fig:trainingDataRepresentations}f). These representations allow incorporating the regional information not only from touching cells but also from close cells resulting in more robust deep learning models. A segmentation network based on the U-Net architecture~\cite{Ronneberger2015}, modified similar to~\cite{Li2019}, is utilized as the backbone of the method. An overview of the proposed method provides Fig~\ref{fig:architecture}.

\subsubsection*{Cell distances and neighbor distances}
The cell distances, as shown in Fig~\ref{fig:trainingDataRepresentations}e, are generated from ground truth data by computing the Euclidean distance transform for each cell independently. Adjacent cells are treated as background in this step and the distance transform is normalized into the range $[0, 1]$. Thus, each pixel of a cell represents the normalized distance to the nearest pixel not belonging to this cell. The cell distance prediction alone is sufficient to obtain seeds for the post-processing. However, a precondition is that the CNN has learned to deal with cell distances of touching cells. By combining cell distances with the novel neighbor distances the erroneous merging of touching cells is prevented.

\begin{figure}
\begin{adjustwidth}{-2.25in}{0in}
\centering
\includegraphics[scale=1.0]{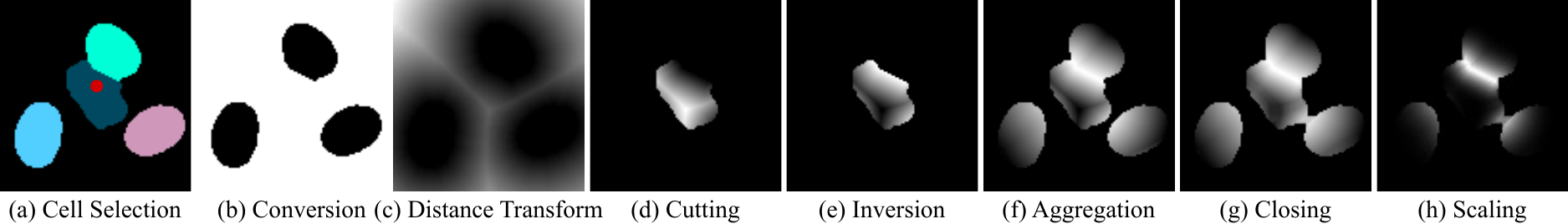}
\caption{\textbf{Main steps of the neighbor distance creation.} After the automated selection of a cell (a), indicated with red, the selected cell and the background are converted to foreground (white in b) while the other cells are converted to background (black in b). Then, the distance transform is calculated (c), cut to the cell region and normalized (d). After inversion (e), the steps are repeated for the remaining cells (f). Finally, the grayscale closed neighbor distances (g) are scaled~(h). Shown is a crop of the Broad Bioimage Benchmark Collection data set BBBC039v1~\cite{Ljosa2012}.}
\label{fig:label-creation}
\end{adjustwidth}
\end{figure}

Fig~\ref{fig:label-creation} shows the generation of the neighbor distances in which each pixel of a cell represents the inverse normalized distance to the nearest pixel of the closest neighboring cell. Therefore, a background-foreground conversion step is applied for each cell independently (Fig~\ref{fig:label-creation}b) and the Euclidean distance transform (Fig~\ref{fig:label-creation}c) is calculated. The distance transform is cut to the cell size and normalized~(Fig~\ref{fig:label-creation}d) followed by an inversion~(Fig~\ref{fig:label-creation}e). The normalization to the range $[0,1]$ is required to suppress neighbor distances for cells without close neighbors. To further reduce the erroneously merging of cells, gaps between close cells are closed by applying a grayscale closing~(Fig~\ref{fig:label-creation}g). Finally, to get a steeper decline within cells, a scaling is applied by taking the closed neighbor distances~(Fig~\ref{fig:label-creation}g) to the power of three~(Fig~\ref{fig:label-creation}h). This confines the neighbor distances to the outer cell area and therefore eases the seed extraction in the post-processing. An advantage of the neighbor distances is that they also provide information in the training process when cells are close but do not touch. This can be seen in Fig~\ref{fig:label-creation}h (bottom right cell and bottom left cell) and is especially advantageous for training data sets with few touching cells providing only little border information in the training process.

\subsubsection*{Robustness of neighbor distances to annotation inconsistencies}
\begin{figure}
\begin{adjustwidth}{-2.25in}{0in}
\centering
\includegraphics[scale=1.0]{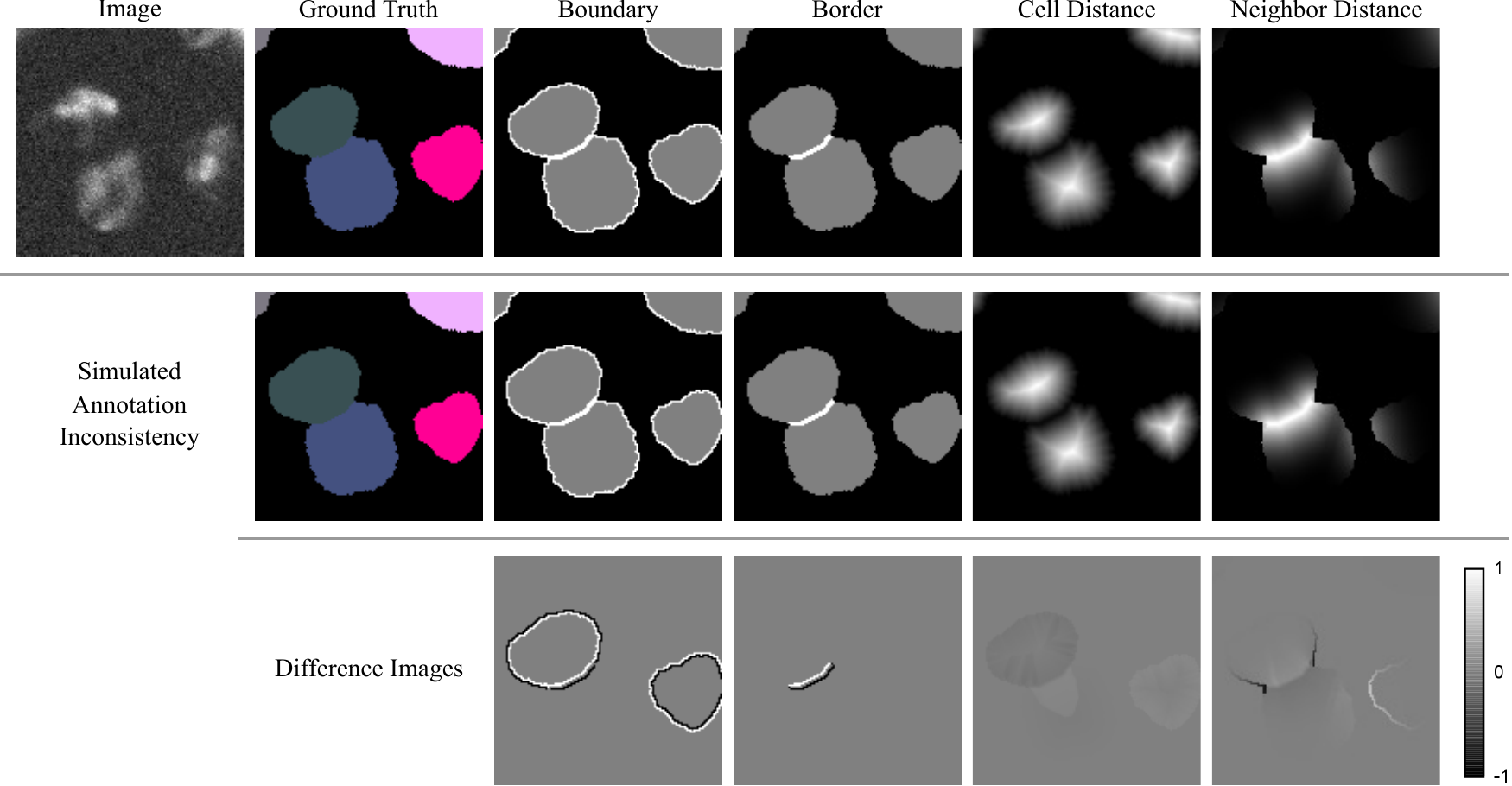}
\caption{\textbf{Robustness of training data representations to annotation inconsistencies.} Small changes in the ground truth, simulated with morphological erosions and dilations, result in different boundaries and borders (first and second row). The difference images between the first row and the second row show that the changes for the distance labels are smoother. Shown is a crop of the Cell Tracking Challenge data set Fluo-N2DH-SIM+ \cite{Ulman2017, Maska2014}.}
\label{fig:trainingDataShift}
\end{adjustwidth}
\end{figure}

Fig~\ref{fig:trainingDataShift} shows that the neighbor distances are more robust to annotation inconsistencies than boundaries and borders, i.e., a cell was morphologically eroded and another cell dilated resulting in masks that only differ in single pixels. The location of the discrete boundaries and borders change, meaning that a prediction of the initial border is considered incorrect and penalized in the training. This makes it difficult to train models well on small data sets. In contrast, the proposed continuous neighbor distance shows a smooth change. Therefore, the influence of annotation inconsistencies on the training process is reduced resulting in a more robust training.

\subsubsection*{Architecture}
The architecture is based on the U-Net architecture~\cite{Ronneberger2015}. Instead of a single decoder path, two parallel decoder paths are used allowing each path to focus on features related to the desired output. In addition, the feature detection in the shared encoder branch of the network is trained using backpropagated information from both decoder branches. The maximum pooling layers are replaced with 2D convolutional layers with stride 2 and kernel size 3. Additionally, batch normalization layers are added. The number of feature maps is doubled from 64 feature maps to a maximum of 1024 in the encoder path and halved in each decoder path correspondingly. To avoid the need of cropping before concatenation, zero padding is applied in the convolutional layers to keep the feature map size consistent. The rectified linear unit activation function is used within the network and a linear activation for the output layers. Fig~\ref{fig:architecture} provides an overview of the architecture, i.e., convolutional and downsampling layers are summarized into~blocks.

\subsubsection*{Watershed post-processing}
\begin{figure}
\begin{adjustwidth}{-2.25in}{0in}
\centering
\includegraphics[scale=1.0]{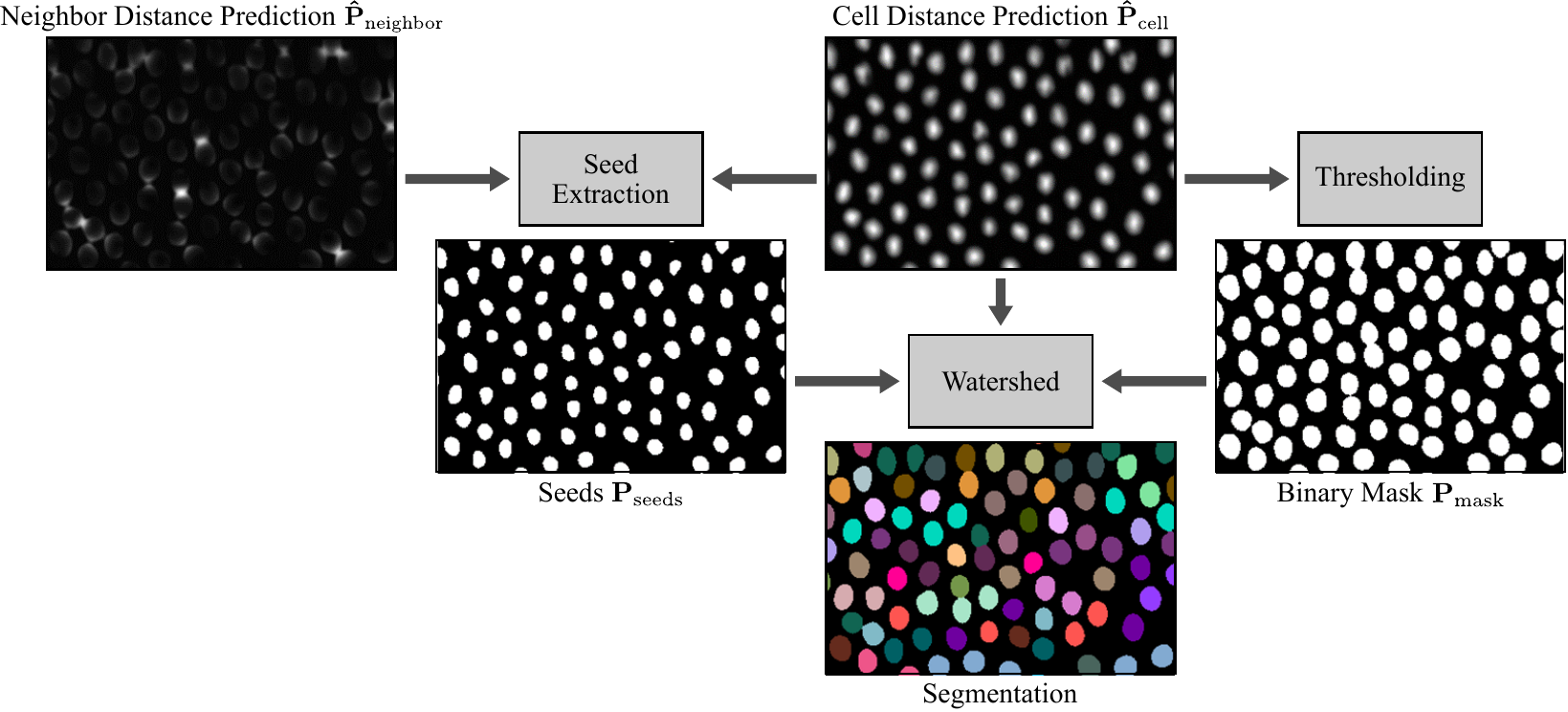}
\caption{\textbf{Overview of the watershed post-processing for segmentation.} The post-processing consists of a threshold-based seed extraction and mask creation, and a watershed. The predictions show a 2D crop of the Cell Tracking Challenge data set Fluo-N3DL-TRIC~\cite{Ulman2017, Maska2014}.}
\label{fig:post-processing}
\end{adjustwidth}
\end{figure}

Fig~\ref{fig:post-processing} shows the main steps of the post-processing. The cell distance prediction $\mathbf{P}_{\text{cell}}$ and the neighbor distance prediction $\mathbf{P}_{\text{neighbor}}$ are smoothed to avoid the erroneous splitting of cells in the seed extraction step:
\begin{align}
	\mathbf{\hat{P}}_{\text{cell}} &= \mathbf{P}_{\text{cell}} *  \mathbf{G}\!\left(\sigma\right) \; ,\\	
	\mathbf{\hat{P}}_{\text{neighbor}} &= \mathbf{P}_{\text{neighbor}} *  \mathbf{G}\!\left(\sigma\right)\; ,
\end{align}
with $ \mathbf{G}\!\left(\sigma\right)$ representing a Gaussian kernel with standard deviation $\sigma$ and $*$ being the convolution operator. Then, the region to flood $\mathbf{P}_{\text{mask}}$ with a seed-based watershed is extracted from the smoothed cell distance prediction by applying a threshold $\varrho_{\text{mask}}$:
\begin{equation}
	\mathbf{P}_{\text{mask}} = \mathbf{\hat{P}}_{\text{cell}} > \varrho_{\text{mask}} \; .
\end{equation}To obtain the seeds, the smoothed and squared neighbor distance prediction is subtracted from the cell prediction and the threshold $\varrho_{\text{seed}}$ is applied:
\begin{equation}
	\mathbf{P}_{\text{seeds}} = \left( \mathbf{\hat{P}}_{\text{cell}} - \mathbf{\hat{P}}_{\text{neighbor}}^2 \right) > \varrho_{\text{seed}}\; .
\end{equation}Depending on the cell size, the squaring can be omitted or replaced by an even steeper function to fine-tune the seed extraction. Seeds with an area smaller than $\num{3}\,\text{px}^2$ are removed. For 3D and 3D+t data, detected merged cells in $z$-direction can be split by increasing the seed extraction threshold $\varrho_{\text{seed}}$ till multiple seeds are found for the merged cells. For the detection of merged cells, a priori knowledge about cell sizes or an outlier detection can be used.

Inspired by the Dual U-Net architecture~\cite{Li2019}, we first attempted to enforce the CNN to predict an additional seed output from the cell distances and the neighbor distances. However, our traditional post-processing provided better results in tests and enables a fine-tuning to cell types not included in the training data. In addition, it simplifies the architecture and the training process.

\subsection*{Cell tracking}
Cell tracking aims to reconstruct the lineage of cells, by linking related cells over time. This linking task is trivial in case of low cell density, error-free segmentation, and high temporal resolution resulting in negligible cell movements between adjacent frames. However, especially for low signal-to-noise-ratio images with touching and dividing cells, fragmented tracks can occur. To re-link fragmented tracks, we match tracks without assigned cells over a short sequence of frames and add a coarse position estimation to the cost function. The proposed algorithm is capable of tracking all segmented cells in an image sequence as well as tracking only a subset, e.g., a selection of manually marked cells.

\subsubsection*{Initialization} 
The tracking algorithm traverses the image sequence $\mathcal{I}=\{\mathbf{I}_t\, |\, 0 \le t \le T\}$ forwards, where $\mathbf{I}_t$ is the image at time point $t$ and $T$ the number of time points. A track is initialized for each segmented object in the first frame. For data sets with marked objects in the first frame, tracks are only initialized for marked objects. It is assumed that the object movement between successive frames is small compared to the overall image size. Therefore, for each tracked object a rectangular region of interest (ROI) is defined as a search space for the same object in the next frame. The initial center of each ROI is set to the median position calculated from the first assigned segmentation mask of each track.

\subsubsection*{Movement estimation} The tracking step consists of a movement estimation followed by a graph-based matching strategy. To estimate the movement of an object, the image frames $\mathbf{I}_t$ and $\mathbf{I}_{t+1}$  are cropped to the object ROI. Then, a phase correlation \cite{Kuglin1975} is calculated between the image crops to estimate a shift between those. The object movement is the shift between the image crops which is given by the position of the maximum peak of the phase correlation. Based on the estimated object movement, the ROI at time point  $t+1$ is adapted for each object individually.

\subsubsection*{Graph-based matching strategy}
All tracks with no successors and their last assigned segmentation mask within time span $\{t-\Delta t, \dots, t\}$ are considered active. Therefore, tracks with missing segmentation masks over at most $\Delta t$ time points can be re-linked. Next, for each active track a set of potential matching candidates is selected based on its ROI at time point $t+1$. Active tracks and potential matching candidates are matched by using an adapted version of the coupled minimum-cost flow algorithm proposed in~\cite{Padfield2011}. The algorithm minimizes the overall cost by selecting edges in the graph with minimal cost subject to a set of constraints. The constraints model flow, appearance/disappearance of objects, and splitting/merging of objects. For an in depth introduction please see~\cite{Padfield2011}.

\begin{figure}
\begin{adjustwidth}{-2.25in}{0in}
\centering
\includegraphics[scale=1.0]{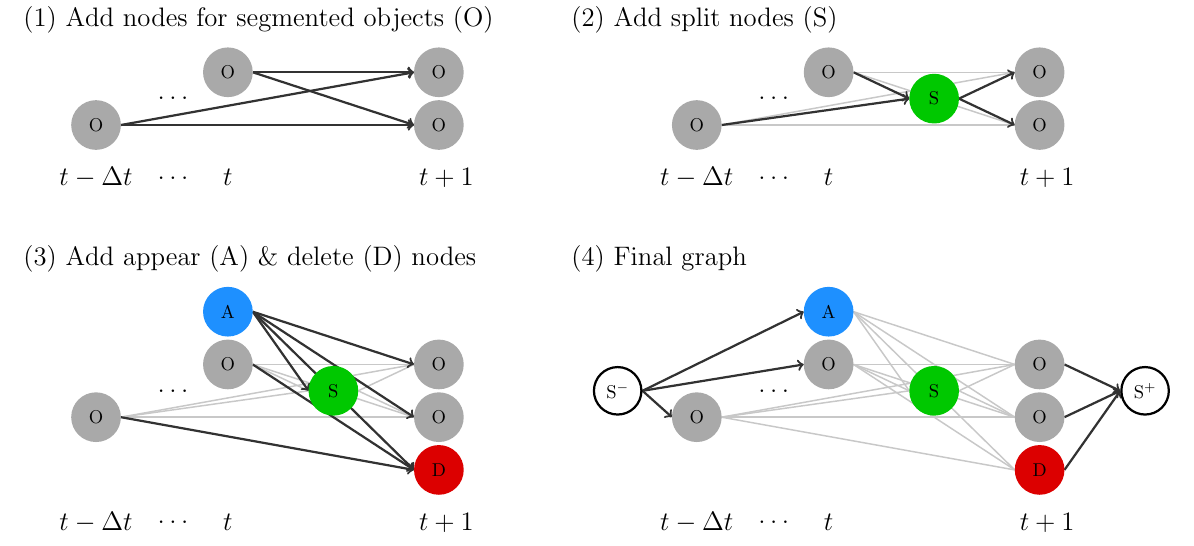}
\caption{\textbf{Graph construction steps exemplary for four segmented objects}. Edges added in a construction step are black, edges added in previous steps are gray. The gray nodes (O) correspond to segmented objects. The segmented objects from $\{t-\Delta t,...,t\}$ are the last matched objects of all active tracks, whereas the segmented objects of $t+1$ are not matched to tracks yet. The blue node models the appearance of objects (A), the red node the disappearance of objects (D), and the green node split events (S). Split event nodes (S) are added for each pair of objects at $t+1$. Therefore, a split event node (S) has exactly two outgoing edges but can have several ingoing edges from object nodes (O). Source ($\text{S}^-$) and sink nodes ($\text{S}^+$) are added for the formulation as coupled minimum cost flow problem.}
\label{fig:mcf_graph}
\end{adjustwidth}
\end{figure}

Fig \ref{fig:mcf_graph} shows our adapted graph structure of the coupled minimum-cost flow algorithm. The following adaptations are applied: the appearance cost of objects is set to 0, as spurious tracks will be filtered out by the subsequent post-processing. This appears to be advantageous in scenarios with the objective to track only a few selected objects. The disappearance cost is set to the length of the largest edge of the ROI instead of using appearance-based features. Therefore, tracks with missing segmentation masks can be assigned to the disappearance node as well. The merging node proposed in~\cite{Padfield2011} is removed, as it only models the merging of two objects per time point. The matching cost $c_{s,n}$ between track $s$ and potential matching candidate $n$ is adapted to: 
\begin{equation}
      c_{s,n}=\left\|\hat{p}^s_{t+1}-p^n_{t+1}\right\|_2,
\end{equation}
where $\hat{p}^s_{t+1}$ is the estimated position of the tracked object $s$ at time point $t+1$ and $p^n_{t+1}$ is the position of the potential matching candidate. The estimated position $\hat{p}^s_{t+1}$ is given by:  
\begin{equation}
    \hat{p}^s_{t+1}=p^s_{t}+d^s_{t,t+1},
\end{equation}
where $d^s_{t,t+1}$ is the estimated shift of the ROI of track $s$ between time points $t$ and $t+1$. $p^s_{t}$ is the position of the tracked object at time point $t$. The cost of split events are computed based on the size and position of the tracked object $s$ and its potential successor candidates $n$ and $k$:
\begin{equation}
     c_{s,(n,k)} = \begin{cases}
     \left\|\hat{p}^s_{t+1}-\frac{1}{2}(p^n_{t+1}+p^k_{t+1})\right\|_2 & \text{if } C_{\text{s}}=1,\\
    \rho & \text{else},\\
     \end{cases}
\end{equation}
where $c_{s,(n,k)}$ are the split costs and $C_{\text{s}}$ the split condition. In practice, we set $\rho$ to ten times the disappearance cost. The split condition $C_{\text{s}}$ is given by:
\begin{equation}
        C_{\text{s}} = \begin{cases} \small
        1 & \text{if } \frac{V^{n}_{t+1}}{V^{k}_{t+1}}\! >\! \alpha, \frac{V^n_{t+1}+V^k_{t+1}}{V^s_{t_{last}}}\! <\! \beta, \left\|p^{n}_{t+1}-p^{k}_{t+1}\right\|_2\! <\! \gamma,\\
        0 & \text{else}.
    \end{cases}
    \label{eq:split}
\end{equation}
$V^{n}_{t+1}, V^{k}_{t+1}$ and $V^{s}_{t_{last}}$ are the sizes of the segmentation masks of successor candidates $n$ and $k$, and of the last assigned object to the track $s$ at time point $t_{last}$, respectively. The successor candidates are sorted so $V^{k}_{t+1} \geq V^{n}_{t+1}$ holds. $\alpha$, $\beta$ and $\gamma$ are hyper-parameters. A possible parametrization of those hyper-parameters is provided in the results section. The split condition ensures that successors are of similar size, have a combined size similar to the size of the predecessor object, and should be reasonably close to each other. 
 
Each active track is only linked to segmented objects overlapping with its ROI, reducing the number of edges in the graph. As all active tracks are added to the graph and not only segmented objects between successive time points, tracks with missing segmentation masks over a short sequence of frames can be linked. A solution of the matching problem is then found using integer linear programming.

For data sets with the aim to track all segmented objects, each non-matched object at time point $t+1$ is initialized as a new track.

\subsubsection*{Post-processing}
In the post-processing step, missing segmentation masks are added by placing masks at the linearly interpolated positions between $t_{last}$ and $t+1$. Furthermore, trajectories of length one without any predecessor and successors are removed.

\section*{Results}
\subsection*{Data set}
\begin{figure}
\centering
\includegraphics[scale=1.0]{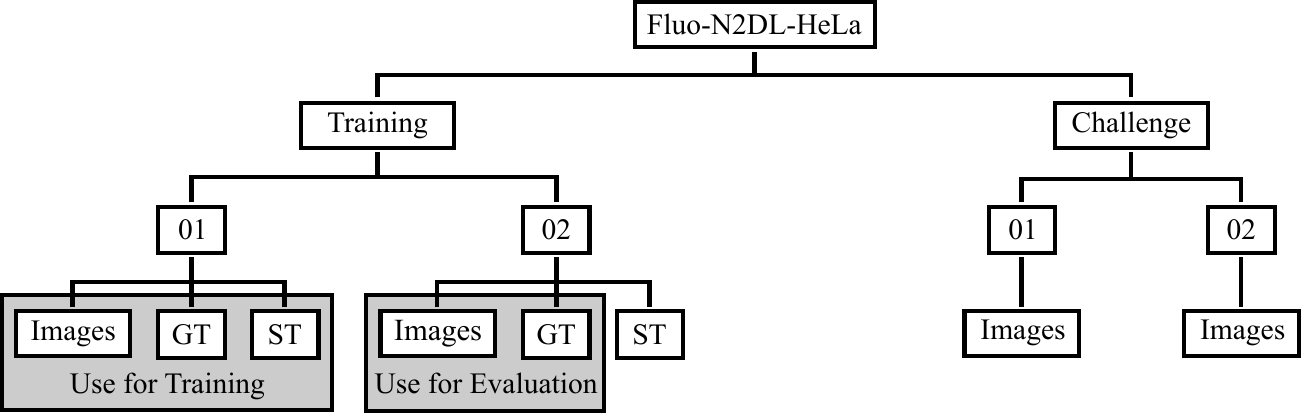}
\caption{\textbf{Cell Tracking Challenge data set structure.} Since no ground truths are publicly available for the challenge sets, the two provided training sets need to be split into a set used for training and a set for evaluation.}
\label{fig:ctc_data_structure}
\end{figure}

We conduct our experiments with data from the Cell Tracking Challenge \cite{Ulman2017, Maska2014}. For each cell type, the provided data sets are split into two training sets with publicly available ground truths, and two challenge sets~(see Fig \ref{fig:ctc_data_structure}). For our experiments, we use selected data from one training set to train models and evaluate on the other. The provided annotations consist of gold truth (GT) instance segmentation masks, interlinked GT cell seeds for cell detection and tracking, and computer-generated instance segmentation masks, referred to as silver truth (ST). The ST annotations, computed from a majority vote of submitted algorithms of former challenge participants, can include segmentation errors. The GT segmentation masks not necessarily include all cells in a frame.

For four data sets, we manually selected segmentation GTs where all cells in a frame are annotated, and STs that do not show obvious segmentation errors. 27~GTs of the data set BF-C2DL-HSC (Mouse hematopoietic stem cells), 15~GTs of the data set BF-C2DL-MuSC (Mouse muscle stem cells), 16~STs of the data set Fluo-N2DL-HeLa (HeLa cells), and 3~GT slices of the 3D data set Fluo-N3DH-CE (\textit{C. elegans} developing embryo) fulfilled our requirements. This heterogeneous data set will be referred to as CTC training set and consists of 268 crops of size $256\,\text{px}\!\times\!256\,\text{px}$ including 52 crops for validation. A difficulty is that the CTC training set contains comparatively few touching cells, whereas in the evaluation the segmentation of touching cells is important, especially for late time points after many cell divisions. Each cell type is evaluated separately using all detection and segmentation GTs of the second set (see Fig~\ref{S1_Fig}).

\subsection*{Evaluation criteria}
For evaluation, we use the performance measures of the Cell Tracking Challenge. The normalized acyclic oriented graph matching measure for detection $\mathrm{DET}$ is used to evaluate object level segmentation errors~\cite{Matula2015}. Pixel level segmentation errors are evaluated with the Jaccard similarity index based measure $\mathrm{SEG}$. The normalized acyclic oriented graph matching measure $\mathrm{TRA}$ is used to evaluate the tracking~\cite{Matula2015}. The overall performances for the Cell Segmentation Benchmark (CSB) and Cell Tracking Benchmark (CTB) are calculated as follows:
\begin{align}
    \mathrm{OP}_{\text{CSB}} &= \num{0.5} \cdot (\mathrm{DET} + \mathrm{SEG}),\\
    \mathrm{OP}_{\text{CTB}} &= \num{0.5} \cdot (\mathrm{SEG} + \mathrm{TRA}).
\end{align}

\subsection*{Parameter selection}
The proposed segmentation method has three adjustable post-processing parameters: the mask threshold $\varrho_{\text{mask}}$, the seed threshold $\varrho_{\text{seed}}$ and the standard deviation $\sigma$ of the Gaussian smoothing. We fix them to: $\varrho_{\text{mask}}=\num{0.09}$, $\varrho_{\text{seed}} = \num{0.5}$, $\sigma = (1.5, 1.5)$ for 2D/2D+t data, and $\sigma = (1.5, 1.5, 0.5)$ for 3D/3D+t data. In practice, a fine-tuning of these parameters is only needed if cells are too small or too large ($\rho_{\text{mask}}$) and if multiple splits or merges occur ($\rho_{\text{seed}}$). For the tracking, we computed cell division and movement statistics from tracking ground truth data and chose the following parameters experimentally: $\Delta t=3$ (dimensionless difference of frames), $\alpha=0.5$, $\beta=1.2$ and $\gamma=2 \cdot\! \sqrt[D]{V^s_{t_{last}}}$ with the number of image dimensions $D \in \{2,3\}$. The ROI is set to $150\,\text{px}\!\times\!150\,\text{px}$ for 2D data sets, and to $(100\,\text{px})^3$ for 3D data sets. For some large 3D+t data sets, e.g., Fluo-N3DL-TRIC and Fluo-N3DL-TRIF of the Cell Tracking Challenge, the ROI is reduced to $(60\,\text{px})^3$. Due to the observed variety of the cell division and movement statistics over the different data sets, we expect improved tracking results by fine-tuning the tracking parameters to each data set individually.

\subsection*{Compared segmentation methods}
The proposed segmentation method is compared with boundary and border prediction methods (Fig~\ref{fig:trainingDataRepresentations}c, Fig~\ref{fig:trainingDataRepresentations}d), adapted borders~\cite{Scherr2018}, the Dual U-Net~\cite{Li2019}, and the J4 method proposed in~\cite{Pena2019}.

For the boundary and border prediction methods, we adapt our proposed architecture and use a single decoder path with a three channel output: background, cell, and boundary/border. Instead of the linear activation, the softmax activation is applied in the output layer. 

For the adapted borders \cite{Scherr2018}, we use our proposed network architecture with two decoder paths. One decoder path is trained to predict binary cell masks (sigmoid activation), the other to predict background, eroded cells and adapted borders (softmax activation).

The Dual U-Net method~\cite{Li2019} uses a similar architecture compared to ours but max-pooling layers and a feature fusion block. Intermediate predictions of discrete boundaries (sigmoid activation) and cell distances (linear activation) are forwarded to the feature fusion block which predicts the final segmentation map (sigmoid activation). We removed in our comparison the dropout layer since none of the other compared methods use dropout.

The last method in our comparison is the J4 method~\cite{Pena2019} which uses $J$ regularization to tackle the class imbalance problem. The J4 method predicts a four channel output: background, cell, touching, and gap. We use the same architecture with softmax activation in the last layer as for the boundary and border method.

Detailed information about the post-processing of the compared methods is provided in the supplementary file S2. Similar to the proposed method, seeds with an area smaller than $\num{3}\,\text{px}^2$ are removed for all methods. Table \ref{tab:border_information} shows the boundary and border information in the CTC training set. The proposed method can utilize more border information in the training process. For the boundary method the most information is resulting from non-touching cells.
\begin{table}[t]
\centering
\footnotesize
\caption{\textbf{Boundary and border information in the CTC training set.} Stated are the ratios of boundary/border pixels to all pixels. For the proposed neighbor distances only pixels with a value greater than \num{0.5} are counted in this comparison. Nevertheless, pixels with smaller neighbor distance values can provide information in the training process as well. For the J4 method \cite{Pena2019}, ratios of the touching and of the gap class are provided.}
\label{tab:border_information}
\setlength{\tabcolsep}{7pt}
\renewcommand{\arraystretch}{1.3}
\begin{tabular}{lccccc}
\hline
 & \textbf{Boundary} & \textbf{Border} & \textbf{Adapted Border}& \textbf{J4} & \textbf{Proposed}\\
\hline
Pixel fraction [\textperthousand] & $10.51$ & $0.11$ & $0.33$ & $0.56$ / $0.96$ & $1.51$\\
\hline
\end{tabular}
\end{table}

\subsection*{Training settings, inference and experimental environment}
For each method, eleven models are trained. This allows to evaluate the robustness of the training. Models are trained with a batch size of $8$ using the Adam optimizer~\cite{Kingma2015} and the learning rate is initialized with \num{8e-4}. After $12$ subsequent epochs without validation loss improvement, the learning rate is multiplied by \num{0.25} till a minimum learning rate of \num{6e-5} is reached. The training is stopped when \num{28} subsequent epochs without improvement occurred or $200$ epochs are reached. To learn the cell distances and the neighbor distances, PyTorch's SmoothL1Loss is used and both losses are added. The loss functions used to train the compared methods are provided in the supplementary file S2. During training the augmentations flipping~(probability: \SI{75}{\percent}), scaling~(\SI{30}{\percent}), rotation~(\SI{30}{\percent}), contrast changing~(\SI{30}{\percent}), blurring~(\SI{30}{\percent}), and noise~(\SI{30}{\percent}) are applied randomly in this order, and the training images are min-max normalized into the range [-1, 1].

For inference, each frame of a time series is min-max normalized independently into the range [-1, 1], whereas the whole volume is normalized for 3D data. The normalized data are processed frame-by-frame with 3D data being processed slice-wise. The CNN model inputs are zero-padded if necessary.

We performed the experiments using a system with two NVIDIA TITAN RTX GPUs, Ubuntu 18.04, and a Intel Core i9-9900K CPU with 64 GB RAM. The methods are implemented in Python and PyTorch is used as deep learning framework. Implementations of the proposed method and of the compared methods are available at \href{https://bitbucket.org/t\_scherr/cell-segmentation-and-tracking/}{https://bitbucket.org/t\_scherr/cell-segmentation-and-tracking/}.

\subsection*{Segmentation results}
\subsubsection*{BF-C2DL-HSC}
\begin{figure}
\begin{adjustwidth}{-2.25in}{0in}
\centering
\includegraphics[scale=1.0]{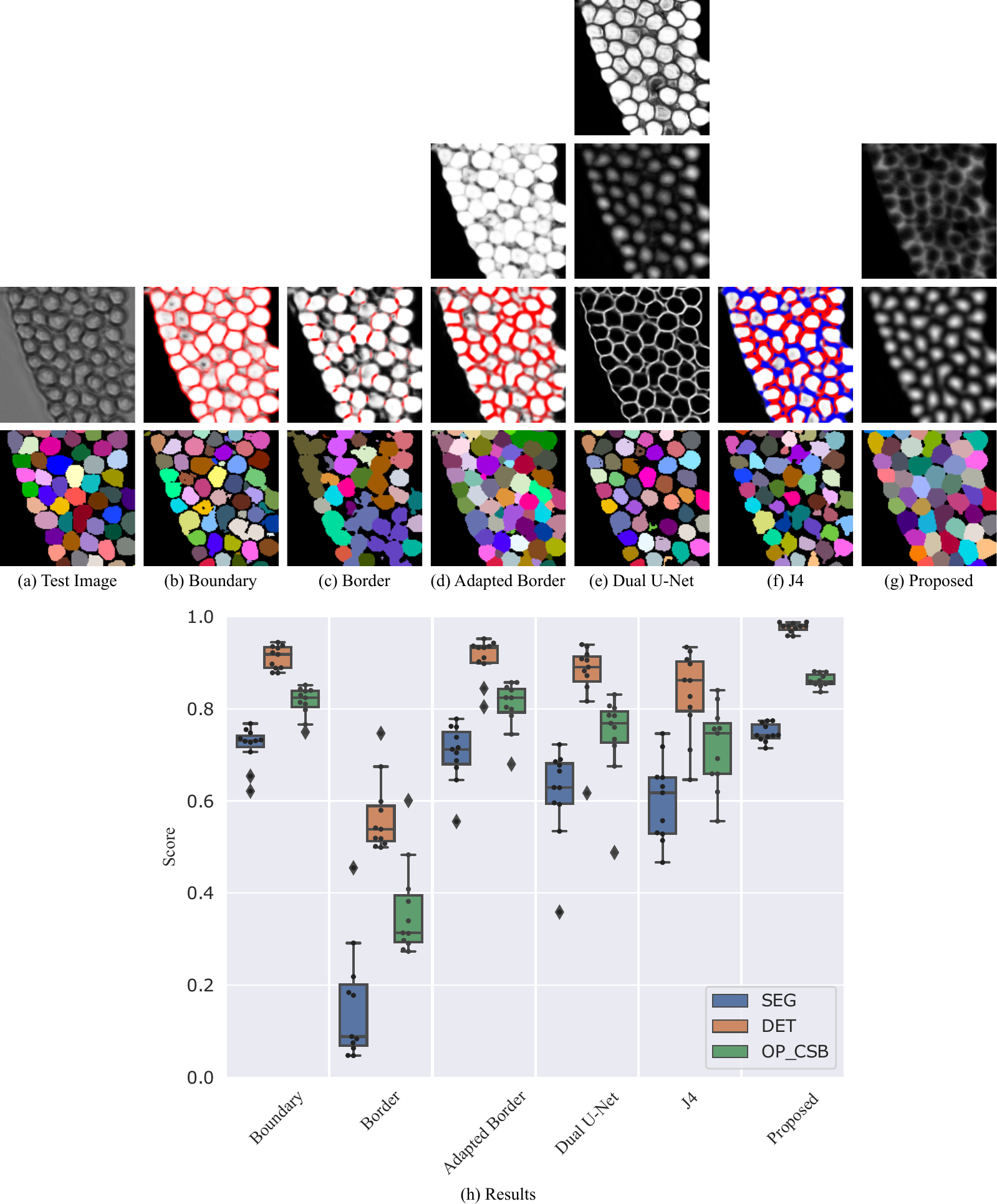}
\caption{\textbf{Segmentation results on the BF-C2DL-HSC test set.} Shown are raw predictions and segmentations of a 140\,px$\times$140\,px test image crop~(a-g, best $\mathrm{OP}_{\text{CSB}}$ models). For multi-channel outputs, channels are color-coded (cell/seed class: white, boundary/border/touching class: red, gap class: blue). The plot at the bottom shows the evaluation on the test set~(h).}
\label{fig:bf-c2dl-hsc-results}
\end{adjustwidth}
\end{figure}

The segmentation results of the Mouse hematopoietic stem cells in Fig~\ref{fig:bf-c2dl-hsc-results} show that the proposed segmentation method provides the best cell detection. The $\mathrm{SEG}$ score, which evaluates pixel level errors, is mainly limited due to the fact that the predicted cells in the proposed method are slightly too large as indicated in Fig~\ref{fig:bf-c2dl-hsc-results}g. These results can be even further improved by fine-tuning the mask threshold. Surprisingly, boundaries can be learned almost as good as adapted borders and better than simple borders. A possible explanation is the small amount of touching cells in the CTC training set which prevents from learning simple borders (Fig~\ref{fig:bf-c2dl-hsc-results}c). The Dual U-Net method suffers from some uncertain regions in the final segmentation map prediction (Fig~\ref{fig:bf-c2dl-hsc-results}e top) resulting in false negatives and split cells. The limitation of the J4 method is that the touching and the gap class are quite similar for this data set. This results in an oversegmentation and imperfect cell shapes since the gap class is considered to be background. The latter limits mainly the $\mathrm{SEG}$ score.

\subsubsection*{Fluo-N3DH-CE}
\begin{figure}
\begin{adjustwidth}{-2.25in}{0in}
\centering
\includegraphics[scale=1.0]{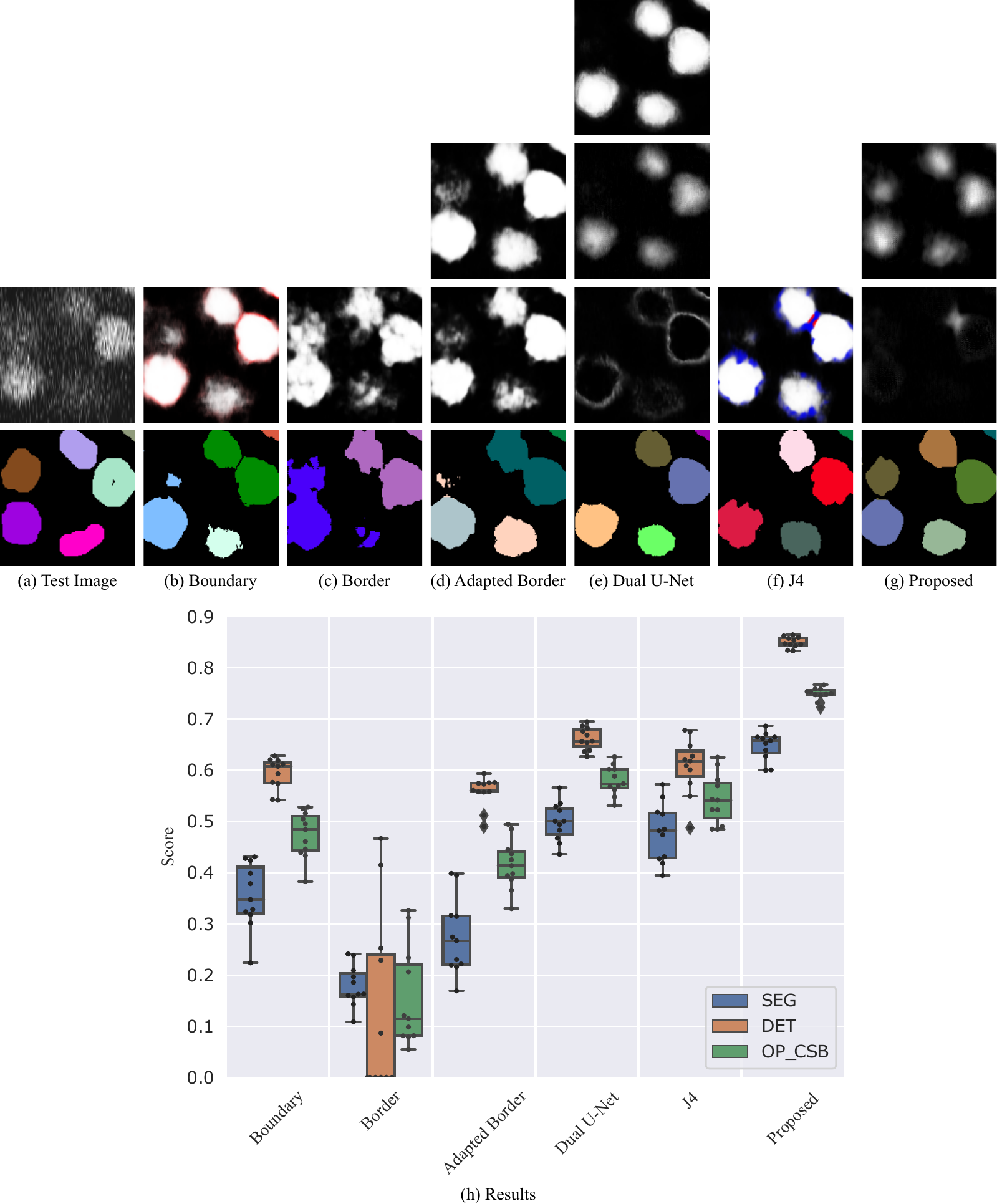}
\caption{\textbf{Segmentation results on the Fluo-N3DH-CE test set.} Shown are raw predictions and segmentations of a 140\,px$\times$140\,px test image crop~(a-g, best $\mathrm{OP}_{\text{CSB}}$ models). For multi-channel outputs, channels are color-coded (cell/seed class: white, boundary/border/touching class: red, gap class: blue). The plot at the bottom shows the evaluation on the test set~(h). Note: this is a 3D data set and the erroneous merging of cells can result from any of the slices a cell appears.}
\label{fig:fluo-n3dh-ce-results}
\end{adjustwidth}
\end{figure}

For the segmentation of the 3D data set Fluo-N3DH-CE, we do not apply the mentioned splitting of cells that are detected as merged. This enables a better comparison of the methods since the almost binary predictions of the other methods do not allow such a simple splitting post-processing. In addition to the 3D nature of this data set, the low signal-to-noise-ratio and the use of only 3 slices of that cell type in the training set makes the segmentation difficult. Again, the proposed method shows the best results whereas boundaries are often unsharp and not closed resulting in merged cells as shown in Fig~\ref{fig:fluo-n3dh-ce-results}. Borders and adapted borders do not appear anymore and cannot be used to split cells. Especially in late frames after many cell divisions, the boundary and border segmentation methods break down and predict only a few very large objects. In contrast, the J4 method shows an improved splitting of touching cells. However, also the J4 method and the Dual U-Net method decrease in segmentation performance in late frames whereas the proposed method still provides a reasonable segmentation.

\subsubsection*{Fluo-N2DL-HeLa}
\begin{figure}
\begin{adjustwidth}{-2.25in}{0in}
\centering
\includegraphics[scale=1.0]{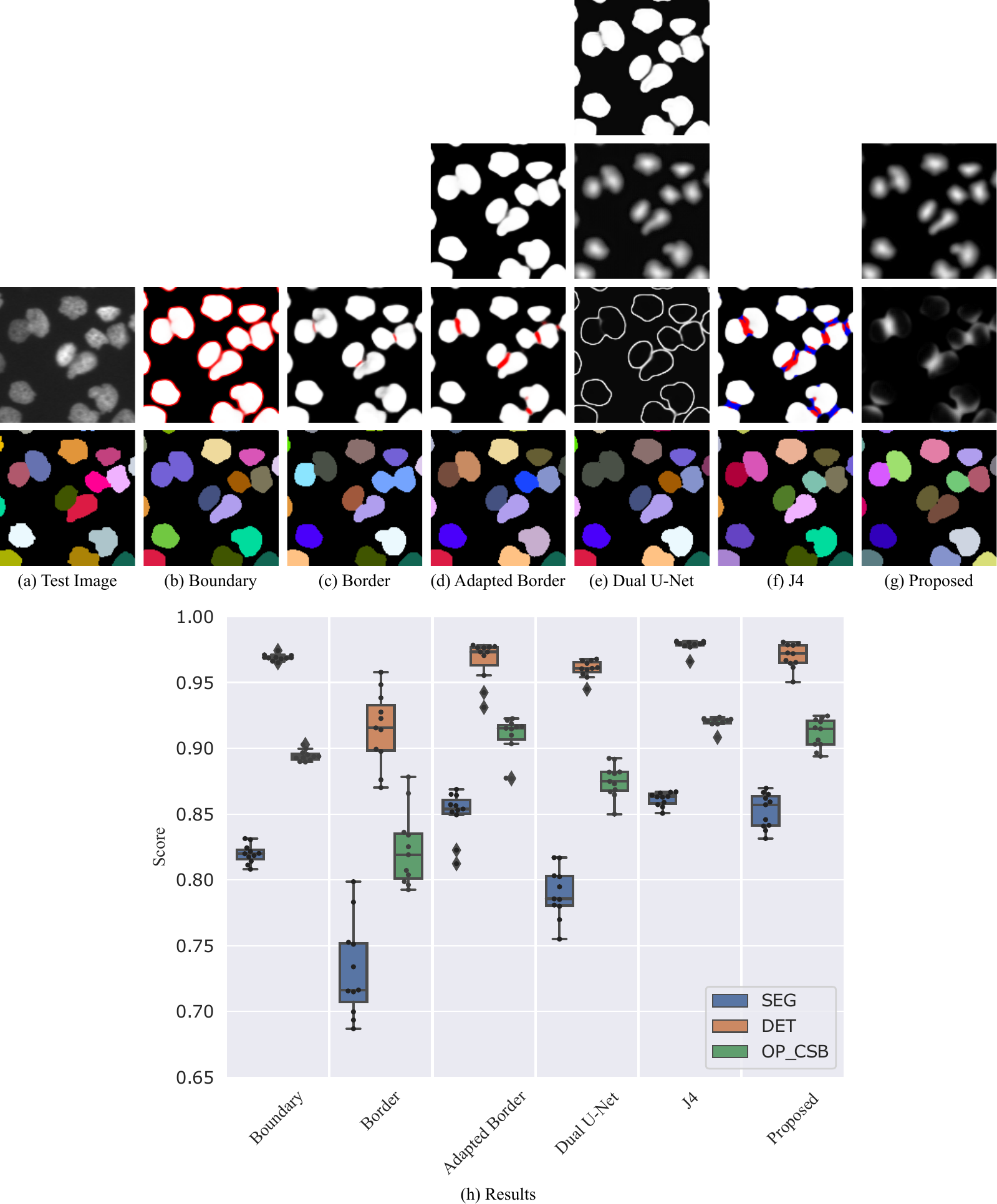}
\caption{\textbf{Segmentation results on the Fluo-N2DL-HeLa test set.} Shown are raw predictions and segmentations of a 140\,px$\times$140\,px test image crop~(a-g, best $\mathrm{OP}_{\text{CSB}}$ models). For multi-channel outputs, channels are color-coded (cell/seed class: white, boundary/border/touching class: red, gap class: blue). The plot at the bottom shows the evaluation on the test set~(h).}
\label{fig:fluo-n2dl-hela-results}
\end{adjustwidth}
\end{figure}

HeLa cells provide the largest quantity of cells from a specific cell type in the CTC training set resulting in the methods performing more similar, as shown in Fig~\ref{fig:fluo-n2dl-hela-results}. For this cell type, the adapted border method shows its advantages over the boundary method by predicting robust borders. The models trained with the J4 method learned to predict and differentiate gap and touching class for this cell type very well resulting in the best performance of all methods. However, the proposed method performs also well for this cell type. The Dual U-Net method suffers from merged objects. This is probably due to non-closed boundaries which induce the merging of cells in the feature fusion layer. Our approach with more robust neighbor distances and a traditional post-processing avoids this.

\subsubsection*{BF-C2DL-MuSC}
\begin{figure}
\begin{adjustwidth}{-2.25in}{0in}
\centering
\includegraphics[scale=1.0]{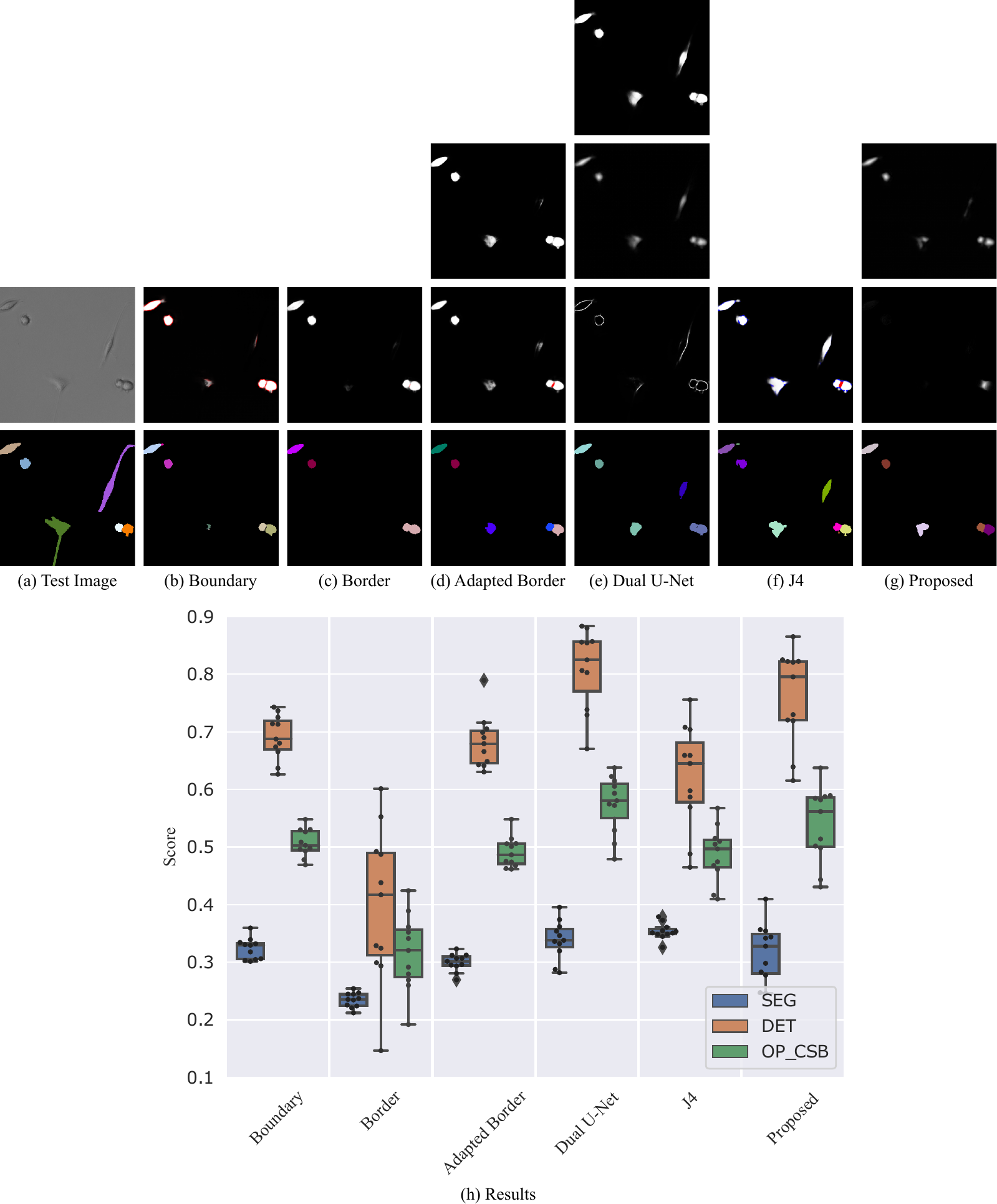}
\caption{\textbf{Segmentation results on the BF-C2DL-MuSC test set.} Shown are raw predictions and segmentations of a 360\,px$\times$360\,px test image crop~(a-g, best $\mathrm{OP}_{\text{CSB}}$ models). For multi-channel outputs, channels are color-coded (cell/seed class: white, boundary/border/touching class: red, gap class: blue). The plot at the bottom shows the evaluation on the test set~(h).}
\label{fig:bf-c2dl-musc-results}
\end{adjustwidth}
\end{figure}

Mouse muscle stem cells are difficult to segment since they change their shape from small roundish objects to elongated objects. Both cell states are shown in Fig~\ref{fig:bf-c2dl-musc-results}. The Dual U-Net method provides the best segmentation of elongated cells, however, roundish cells are sometimes merged. Nevertheless, the better segmentation of the elongated cells compensates this. The J4 method in contrast suffers from oversegmentation on this cell type, resulting in lower scores. As for the data set BF-C2DL-HSC, the proposed method can handle the small roundish cells well resulting in the second best method for this cell type. The segmentation problem of the elongated cells can be solved using the training data of both BF-C2DL-MuSC training data subsets. This is shown in the next section.

\subsection*{Cell Tracking Challenge}
For our submission to the 5th IEEE ISBI 2020 Cell Tracking Challenge as team KIT-Sch-GE, we combined our segmentation method with our adapted tracking approach. We selected a training data set similarly to the CTC training set. The data set consists of 997 crops of size $256\,\text{px}\!\times\!256\,\text{px}$ of carefully selected Cell Tracking Challenge data, CBIA HL60 cell line data~\cite{Svoboda2009}, BBBC038 drosophila images~\cite{Caicedo2019}, and generated semi-synthetic data~\cite{Stegmaier2017, Stegmaier2016}. A more detailed description of the data set, data set specific segmentation and tracking parameters, and executables can be found on the challenge website. For our submission we manually selected a segmentation model from three trained models. To avoid issues with the TRA measure, frames without any tracked object are replaced by the tracking result of the temporally closest frame. For the Fluo-N3DH-CE data set cells are split if their volume is bigger than $\frac{4}{3}$ times the mean object volume at that time point by iteratively increasing the seed extraction threshold $\varrho_{\text{seed}}$.

\subsubsection*{Cell Segmentation Benchmark}
\begin{figure}
\centering
\includegraphics[scale=1.0]{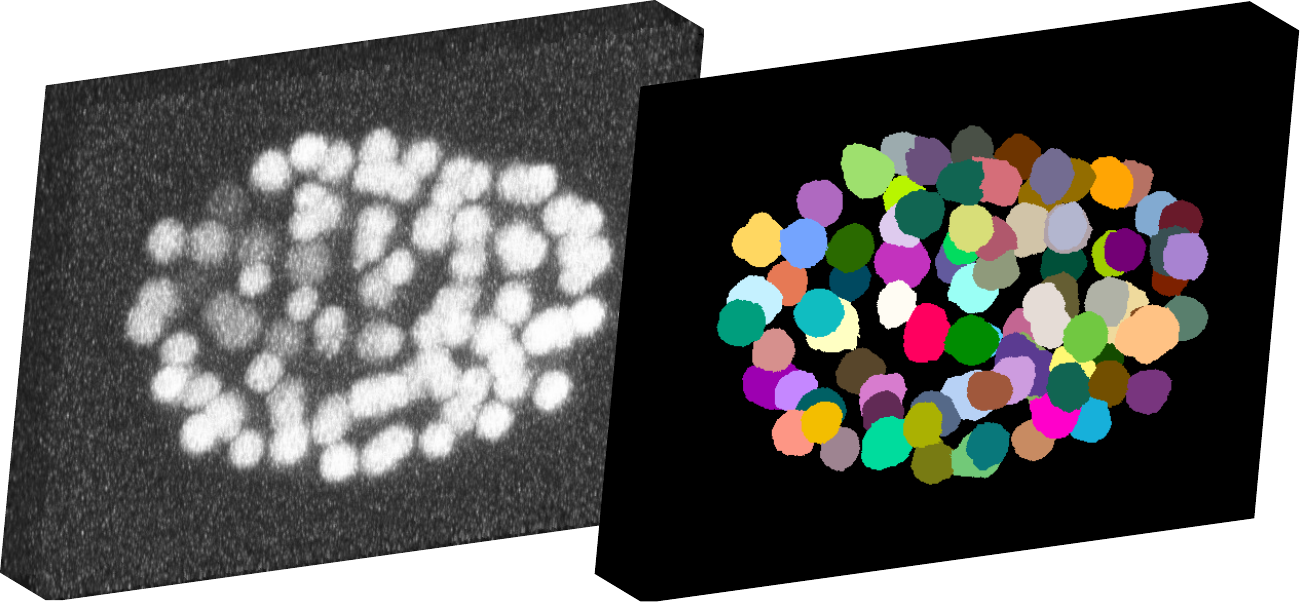}
\caption{\textbf{Segmentation result of the Fluo-N3DH-CE challenge data.} The maximum intensity projection of the raw data (left) and of the segmentation (right) show that cells can be segmented well even on this challenging data set.}
\label{fig:results-ce}
\end{figure}

In the Cell Segmentation Benchmark, we achieved eight top three rankings, including two first places, and three fourth places, all of them with the same model (see Table~\ref{tab:results_ctc}). The performances of the highlighted data sets with no or almost no training data used imply a good utilization of training data and a good generalization ability of our model. A comparison of the scores for the data set BF-C2DL-MuSC on the CSB benchmark and the previous experiment in Fig~\ref{fig:bf-c2dl-musc-results} shows an improved performance. We assume this is due to the larger amount of elongated cells in the training data. Furthermore, the results show that our proposed 2D segmentation with 3D post-processing approach performs well on 3D data. An exemplary segmentation result is shown in Fig~\ref{fig:results-ce}.
\begin{table}[t]
\begin{adjustwidth}{-2.25in}{0in}
\centering
\footnotesize
\caption{\textbf{Cell Tracking Benchmark and Cell Segmentation Benchmark results (5th edition).} Top 3 rankings in the overall performances $\mathrm{OP}_{\text{CSB}}$ and $\mathrm{OP}_{\text{CTB}}$ are written in bold. The corresponding \href{http://celltrackingchallenge.net/files/leaderboards/CTB/2020-04-03.png}{Cell Tracking Benchmark} and \href{http://celltrackingchallenge.net/files/leaderboards/CSB/2020-04-03.png}{Cell Segmentation Benchmark} leaderboards are available on the Cell Tracking Challenge website.}
\label{tab:results_ctc}
\setlength{\tabcolsep}{7pt}
\renewcommand{\arraystretch}{1.3}
\begin{tabular}{lcccccccc}
\hline
\textbf{Data Set} & $\mathrm{\mathbf{SEG}}$ & $\mathrm{\mathbf{DET}}$ & $\mathrm{\mathbf{TRA}}$ & $\mathrm{\mathbf{OP}}_{\textbf{CSB}}$ & $\mathrm{\mathbf{OP}}_{\textbf{CTB}}$ & \textbf{Ranking} $\mathrm{\mathbf{OP}}_{\textbf{CSB}}$ & \textbf{Ranking} $\mathrm{\mathbf{OP}}_{\textbf{CTB}}$\\
\hline
\textbf{BF-C2DL-HSC} & $\mathbf{0.750}$ & $\mathbf{0.974}$ & $\mathbf{0.929}$ & $\mathbf{0.862}$ & $\mathbf{0.840}$ & \textbf{2nd} & \textbf{3rd} \\
\textbf{BF-C2DL-MuSC} & $\mathbf{0.702}$ & $\mathbf{0.977}$ & $\mathbf{0.967}$ & $\mathbf{0.839}$ & $\mathbf{0.835}$ & \textbf{1st} & \textbf{1st} \\
Fluo-C3DH-H157\textsuperscript{\textdagger} & \num{0.789} & \num{0.949} & \num{0.948} & \num{0.869} & \num{0.869} & 4th & 4th \\
\textbf{Fluo-C3DL-MDA231}\textsuperscript{\textdagger\textdagger} & $\mathbf{0.616}$ & $\mathbf{0.851}$ & $\mathbf{0.820}$ & $\mathbf{0.733}$ & $\mathbf{0.718}$ & \textbf{3rd} & \textbf{3rd} \\
Fluo-N2DH-GOWT1 & \num{0.828} & \num{0.950} & \num{0.949} & \num{0.889} & \num{0.889} & 14th & 12th \\
\textbf{Fluo-N2DL-HeLa} & $\mathbf{0.895}$ & $\mathbf{0.992}$ & $\mathbf{0.989}$ & $\mathbf{0.944}$ & $\mathbf{0.942}$ & \textbf{3rd} & \textbf{3rd} \\
\textbf{Fluo-N3DH-CE}\textsuperscript{\textdagger\textdagger} & $\mathbf{0.729}$ & $\mathbf{0.930}$ & $\mathbf{0.886}$ & $\mathbf{0.830}$ & $\mathbf{0.808}$  & \textbf{1st} & \textbf{1st} \\
\textbf{Fluo-N3DH-CHO} & $\mathbf{0.871}$ & $\mathbf{0.945}$ & $\mathbf{0.948}$ & $\mathbf{0.908}$ & $\mathbf{0.909}$ & \textbf{3rd} & \textbf{3rd} \\
Fluo-N3DL-DRO & \num{0.562} & \num{0.761} & - & \num{0.661} & - & 4th & - \\
\textbf{Fluo-N3DL-TRIC}\textsuperscript{\textdagger} & $\mathbf{0.821}$\,/\,$\mathbf{0.766}^{*}$ & $\mathbf{0.961}$ & $\mathbf{0.809}$ & $\mathbf{0.891}$ & $\mathbf{0.787}$ & \textbf{2nd} & \textbf{2nd} \\
\textbf{Fluo-N3DL-TRIF}\textsuperscript{\textdagger} & $\mathbf{0.601}$\,/\,$\mathbf{0.573}^{*}$ & $\mathbf{0.926}$ & $\mathbf{0.788}$ & $\mathbf{0.763}$ & $\mathbf{0.680}$ & \textbf{3rd} & \textbf{3rd} \\
Fluo-N2DH-SIM+ & \num{0.800} & \num{0.949} & \num{0.945} & \num{0.875} & \num{0.873} & 9th & 7th \\
\textbf{Fluo-N3DH-SIM+} & $\mathbf{0.668}$ & $\mathbf{0.937}$ & $\mathbf{0.933}$ & $\mathbf{0.802}$ & $\mathbf{0.800}$ & \textbf{4th} & \textbf{2nd} \\
\hline
\multicolumn{9}{p{251pt}}{\textsuperscript{\textdagger} No data of that cell type used to train the segmentation model.}\\
\multicolumn{9}{p{251pt}}{\textsuperscript{\textdagger\textdagger} $\leq 5$ slices of that cell type used to train the segmentation model.}\\
\multicolumn{9}{p{495pt}}{\textsuperscript{*} Two scores for each benchmark (CSB/CTB) due to a different treatment of additionally detected and segmented cells.}\\
\end{tabular}
\end{adjustwidth}
\end{table}

\subsubsection*{Cell Tracking Benchmark}
\begin{figure}
\centering
\includegraphics[scale=1.0]{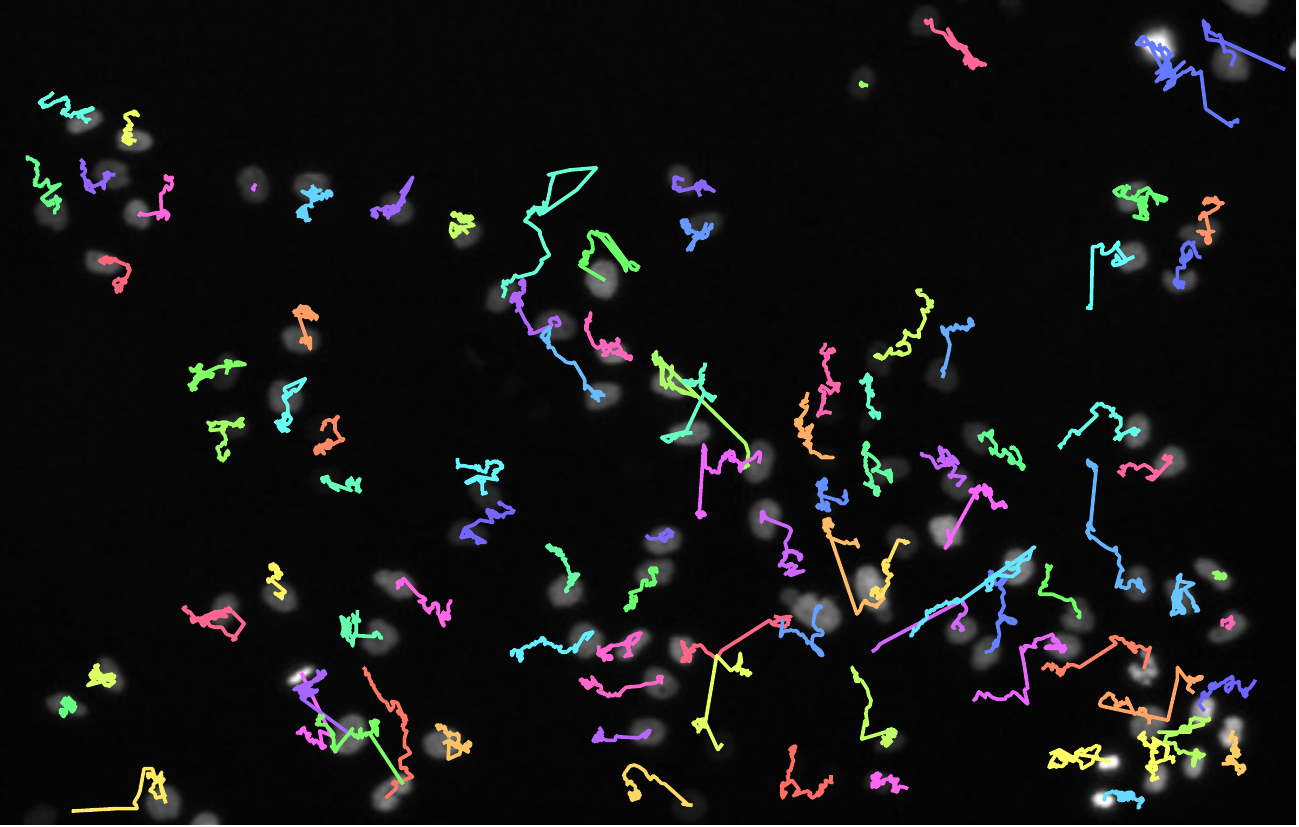}
\caption{\textbf{Tracking results on the Fluo-N2DL-HeLa challenge data.} The first raw image is overlaid with the tracks starting in the first frame. For better visibility, tracks starting in later frames are excluded.}
\label{fig:results-track-hela}
\end{figure}

In the Cell Tracking Benchmark, we achieved nine top three rankings, including two first places, and a fourth place~(see Table~\ref{tab:results_ctc}). Exemplarily, some tracking results of our approach are shown in Fig~\ref{fig:results-track-hela}. Some tracks show jumps, visible as long straight lines, possibly due to some remaining linking errors in our adapted tracking approach. However, none of the competing tracking approaches yields perfect tracking results for all cells. The multiple top performances in the Cell Tracking Benchmark show that our tracking approach, which combines a movement estimation and a graph-based matching strategy, belongs to the best performing approaches.

\section*{Discussion}
Trained on a data set with few touching samples, our proposed segmentation method outperforms all compared methods for at least three of the four cell types evaluated. This is due to the fact that the neighbor distances enable our method to learn from close cells which results in additional information in the training process (see Table \ref{tab:border_information}) and the fact that this information can be easily combined with the cell distances. The differences between the segmentation results of both discrete border methods show how important the utilization of border information is. We want to emphasize that for the proposed method the seed and the mask thresholds can be adjusted for each cell type and for each trained model separately. This improves the segmentation results shown in Fig~\ref{fig:bf-c2dl-hsc-results}~-~\ref{fig:bf-c2dl-musc-results}. The other discrete methods do not allow to do so since the needed sigmoid or softmax activation functions prevent a major fine-tuning of the post-processing. However, to allow a better comparison, we fixed the post-processing parameters of our proposed method in our experiments. The results of the J4 method on the HeLa data set show that specialized loss functions work very well, at least for the dominating cell type in the training set. So far, our approach only uses standard loss functions.

Our successful participation in the Cell Segmentation Benchmark and the Cell Tracking Benchmark show that our proposed tracking by detection method yields excellent results in cell segmentation and cell tracking. Especially the success on data sets with only little or very sparse annotated training data, i.e., data with only very few cells in a frame annotated, shows the advantages of our method.

\section*{Conclusion}
\label{sec:conclusion}
The segmentation and tracking of touching and dividing cells of different types is a challenging task. In this work, a new cell segmentation method using a combination of cell distances and neighbor distances is proposed. The segmentation method utilizes information from touching and close cells in the training process. Therefore, it shows an improved generalization ability for cell types underrepresented or absent in the training data set compared to border and boundary prediction methods. This advantage enables to segment even cell types with no or almost no annotated training data available. Our success in the Cell Segmentation Benchmark emphasizes the strengths of our segmentation method. Our adapted tracking algorithm, which uses a movement estimation with a graph-based matching strategy, can handle cell divisions and missing segmentation masks over a short sequence of frames. The combination of the tracking with our proposed segmentation method resulted in top performances at the Cell Tracking Benchmark.

As future research, we plan to further improve the segmentation performance using a larger and on ImageNet pre-trained encoder or mixed convolution blocks~\cite{Huang2020}, test-time augmentation~\cite{Moshkov2020}, and the synthetic generation of new training samples\cite{Hollandi2019}. In addition, studies about how cell features, e.g., size, shape and texture, influence the generalization ability to new cell types are needed. A long-term goal is to develop a user-friendly-software for the segmentation and tracking of a large variety of cell types using a well-trained segmentation model. Including tunable post-processing parameters facilitates an adaptation of the cell and neighbor distances to new data.

\section*{Author contributions}
Conceived and designed the experiments: TS KL MB RM. Performed the experiments: TS KL. Analyzed the data: TS KL. Contributed reagents/materials/analysis tools: TS KL MB RM. Wrote the paper: TS KL MB RM.

\section*{Funding}
We are grateful for funding by the Helmholtz Association in the programs BioInterfaces in Technology and Medicine (TS, RM) and the Helmholtz Information \& Data Science School for Health (KL, RM). The funders had no role in study design, data collection and analysis, decision to publish, or preparation of the manuscript.

\section*{Competing interests}
The authors have declared that no competing interests exist.

\section*{Acknowledgments}
The authors would like to thank Andreas Bartschat for proofreading, the organizers of the Cell Tracking Challenge, and all data set providers.

\cleardoublepage
\thispagestyle{empty}
\renewcommand\thefigure{S1} 
\begin{figure}[h]
\begin{adjustwidth}{-2.25in}{0in}
\centering
\includegraphics[scale=1.0]{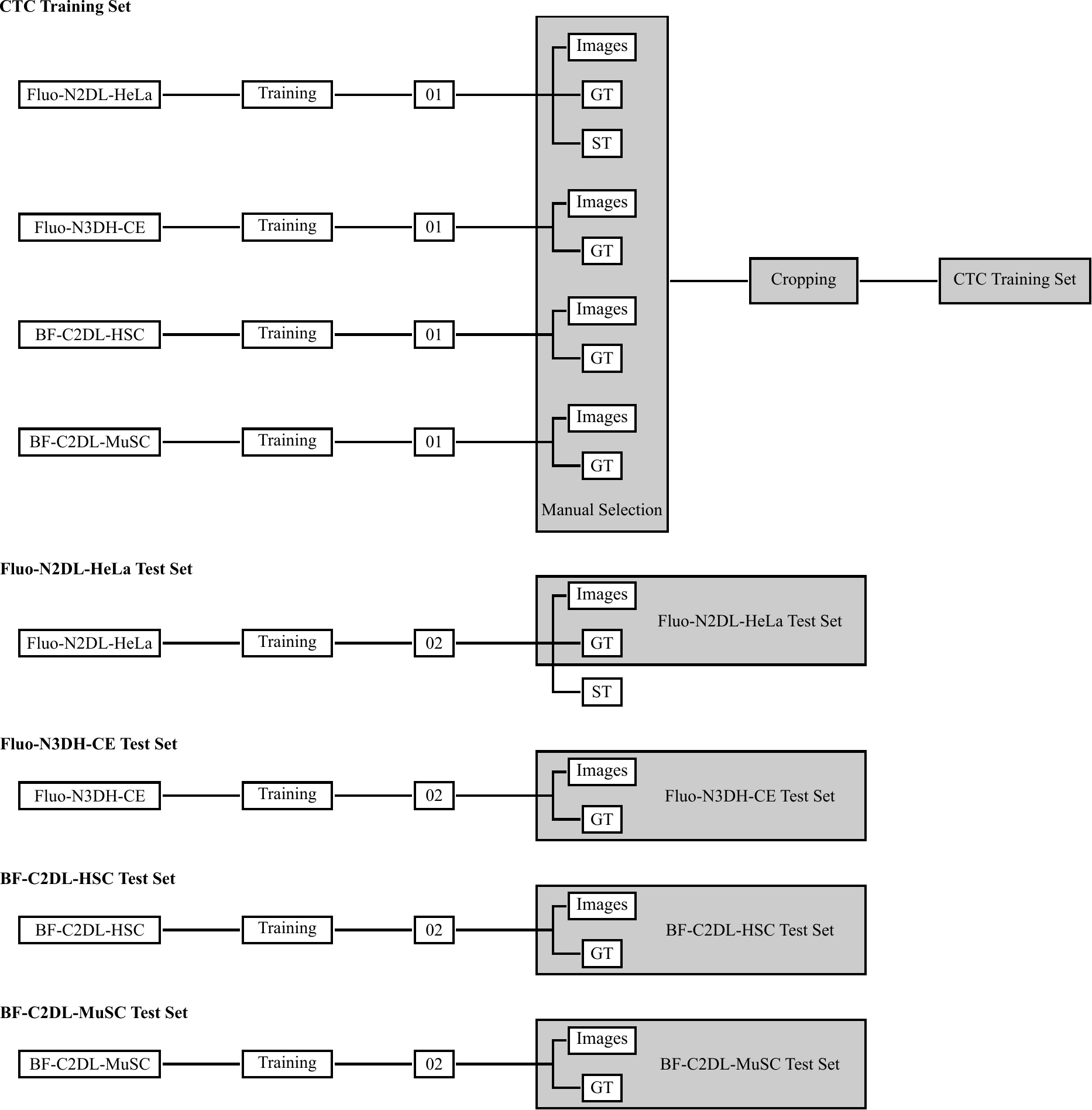}
\caption{\textbf{Training and test data set collection.} The ground truths of the challenge sets from the Cell Tracking Challenge are not publicly available. Thus, the two training data sets are split into a training set and cell type specific test sets for our segmentation experiments. For the training data set only fully annotated segmentation GTs and good quality STs can be used to train models well. For evaluation, all segmentation and detection GTs can be used.}
\label{S1_Fig}
\end{adjustwidth}
\end{figure}
\cleardoublepage



\section*{Supplementary material}

# Cell segmentation and tracking using CNN-based distance predictions and a graph-based matching strategy

## Supplementary file S2: post-processing and loss functions of the compared methods

Tim Scherr[1]*, Katharina Löffler [1,2], Moritz Böhland[1], Ralf Mikut[1]

**1** Institute for Automation and Applied Informatics, Karlsruhe Institute of Technology, Eggenstein-Leopoldshafen, Germany, **2** Institute of Biological and Chemical Systems – Biological Information Processing, Karlsruhe Institute of Technology, Eggenstein-Leopoldshafen, Germany

* tim.scherr@kit.edu

This supplementary file describes the post-processing of the compared methods in detail. The compared methods and our proposed method are based on a seed and mask extraction which are used for a watershed segmentation.

## Boundary method

The boundary method (BM) predicts a three channel output (background, cell, boundary). Activation function is the softmax activation and a weighted sum of Dice loss (channel-wise) and cross entropy loss is used to train this method. For the watershed segmentation binary masks $\mathbf{P}_{\text{BM,mask}}$ and seeds $\mathbf{P}_{\text{BM,seeds}}$ are extracted from the single channels $\mathbf{P}_{\text{BM}}(x)$ with $x \in \{\text{background, cell, boundary}\}$:

$$p_{\text{BM,mask}}^{ij} = \begin{cases} 1 & \text{if } \arg\max_{x} p_{\text{BM}}^{ij}(x) = \text{cell}, \\ 0 & \text{else}, \end{cases} \tag{s1}$$

$$\mathbf{P}_{\text{BM,seeds}} = \mathbf{P}_{\text{BM}}(\text{cell}) \cdot \left[1 - \mathbf{P}_{\text{BM}}(\text{boundary})\right] > 0.5. \tag{s2}$$

$p_{\text{BM,mask}}^{ij}$ represents the value of $\mathbf{P}_{\text{BM,mask}}$ at pixel $(i, j)$. The resulting mask is equal to one at pixels with maximum prediction in the cell channel. For 3D data, Eq. s1 needs to be adapted to the voxel coordinates.

## Border method

The border method (BoM) predicts a three channel output (background, cell, border). Activation function is the softmax activation and a weighted sum of Dice loss (channel-wise) and cross entropy loss is used to train this method. For the watershed segmentation binary masks $\mathbf{P}_{\text{BoM,mask}}$ and seeds $\mathbf{P}_{\text{BoM,seeds}}$ are extracted from the single channels $\mathbf{P}_{\text{BoM}}(x)$ with $x \in \{\text{background, cell, border}\}$:

$$p_{\text{BoM,mask}}^{ij} = \begin{cases} 1 & \text{if } \arg\max_{x} p_{\text{BM}}^{ij}(x) = \text{cell}, \\ 0 & \text{else}, \end{cases} \tag{s3}$$

$$\mathbf{P}_{\text{BoM,seeds}} = \mathbf{P}_{\text{BoM}}(\text{cell}) \cdot \left[1 - \mathbf{P}_{\text{BoM}}(\text{border})\right] > 0.5. \tag{s4}$$

$p_{\text{BoM,mask}}^{ij}$ represents the value of $\mathbf{P}_{\text{BoM,mask}}$ at pixel $(i, j)$. The resulting mask is equal to one at pixels with maximum prediction in the cell channel. For 3D data, Eq. s3 needs to be adapted to the voxel coordinates.

## Adapted border method

The adapted border method (ABoM) predicts a single channel output (cell) used for the mask creation and a three channel output (background, eroded cell, adapted border) used for seed detection. The sigmoid activation function is applied to the single channel cell output and the softmax activation to the three channel output. A weighted sum of Dice loss and cross entropy loss is used to train both outputs. The total loss function is the sum of the loss functions of each output. For the watershed segmentation binary masks $\mathbf{P}_{\text{ABoM,mask}}$ and seeds $\mathbf{P}_{\text{ABoM,seeds}}$ are extracted from the single channels $\mathbf{P}_{\text{ABoM}}(x)$ with $x \in \{\text{cell, background, eroded cell, adapted border}\}$:

$$\mathbf{P}_{\text{ABoM,mask}} = \mathbf{P}_{\text{ABoM}}(\text{cell}) > 0.5, \tag{s5}$$

$$\mathbf{P}_{\text{ABoM,seeds}} = \mathbf{P}_{\text{ABoM}}(\text{eroded cell}) \cdot \left[1 - \mathbf{P}_{\text{ABoM}}(\text{adapted border})\right] > 0.5. \tag{s6}$$

## Dual U-Net method

The Dual-U-Net method predicts a single channel cell distance output (linear activation), a single channel boundary output (sigmoid activation) and a single channel segmentation map output (sigmoid activation). To learn the cell distances a mean square error loss is used, and for each other output a weighted sum of Dice loss and cross entropy loss. The total loss function is a weighted sum of the three losses. The weighting is needed to avoid that only one loss is minimized and was adjusted to the CTC training set.

The intermediate cell distance and boundary outputs are only needed in the training process. For inference, only the segmentation output based on the intermediate predictions is used. In the corresponding publication, no specific post-processing is mentioned. However, we applied a simple morphological erosion after thresholding the segmentation map which reduces the merging of cells in some cases. To get the initial cell shapes we applied a watershed segmentation to the thresholded segmentation map with the eroded cells as seeds.

## J4 method

The J4 method predicts a four channel output (background, cell, touching, gap). As activation function softmax activation is used. To train the network, a weighted sum of cross entropy loss and $J$ regularization loss is used as loss function. Instead of using the cross entropy loss definition of the corresponding publication, we used a PyTorch implementation which is numerically more stable. Since the $J$ regularization loss was much larger than the cross entropy loss on our data (regardless of the cross entropy implementation) we scaled it in a way that the $J$ regularizaton loss was the dominating loss part after minimizing the cross entropy loss, as suggested in the corresponding publication.

For the watershed segmentation binary masks $\mathbf{P}_{\text{J4,mask}}$ and seeds $\mathbf{P}_{\text{J4,seeds}}$ are extracted from the single channels $\mathbf{P}_{\text{J4}}(x)$ with $x \in \{\text{background, cell, touching, gap}\}$

$$p_{\text{J4,mask}}^{ij} = \begin{cases} 1 & \text{if } \arg\max_{x} p_{\text{J4}}^{ij}(x) = \text{cell}, \\ 1 & \text{if } \arg\max_{x} p_{\text{J4}}^{ij}(x) = \text{touching}, \\ 0 & \text{else}, \end{cases} \tag{s7}$$

$$\mathbf{P}_{\text{J4,seeds}} = \mathbf{P}_{\text{J4}}(\text{cell}) \cdot \left[1 - \mathbf{P}_{\text{J4}}(\text{touching})\right] \cdot \left[1 - \mathbf{P}_{\text{J4}}(\text{gap})\right] > 0.5. \tag{s8}$$

$p_{\text{J4,mask}}^{ij}$ represents the value of $\mathbf{P}_{\text{J4,mask}}$ at pixel $(i, j)$. The resulting mask is equal to one at pixels with maximum prediction in the cell channel or in the touching channel (touching class can only occur within cells, gaps within background). For 3D data, Eq. s7 needs to be adapted to the voxel coordinates.

We also tested to just use the arg max of the cell class but this resulted in some additional merged cells. Since merged cells are penalized stronger by the metric than false positives/splits the method benefits from our post-processing in our comparison.



\end{document}